\documentclass{article}
\usepackage{ijcai26}
\newif\ifanonymized
\anonymizedfalse
\newif\ifreview
\reviewfalse
\usepackage{times}
\usepackage{soul}
\usepackage{url}
\usepackage{xurl}
\usepackage[hidelinks]{hyperref}
\usepackage[utf8]{inputenc}
\usepackage[small]{caption}
\usepackage{graphicx}
\usepackage{amsmath}
\usepackage{amsthm}
\usepackage{amssymb}
\usepackage{booktabs}
\usepackage{multirow}
\usepackage{enumitem}
\usepackage{algorithm}
\usepackage{algorithmic}
\usepackage[switch]{lineno}
\ifreview
\linenumbers
\fi
\title{OrthKD: Extracting Generalized Clinical Knowledge from Heterogeneous Teachers for Lightweight Deployment}

\ifanonymized
\author{
    Author Name
    \affiliations
    Affiliation
    \emails
    email@example.com
}
\else
\author{
Yi Xu\thanks{Equal contribution.}\thanks{Corresponding author.}
\and
Cheng Chen\footnotemark[1]
\and
Mufan Cao
\affiliations
Tongji University, Shanghai, China
\emails
\{2351441, 2351445, 2352738\}@tongji.edu.cn
}
\fi

\begin{document}

\maketitle

\begin{abstract}
Deploying diabetic retinopathy (DR) screening models in primary care requires edge-efficient systems that remain accurate, safe, and reliable under domain shift. Multi-teacher knowledge distillation (KD) is a natural compression strategy, but existing approaches largely assume that all teachers provide equally trustworthy supervision. In our setting, this assumption fails: a strong CNN teacher (EfficientNet-B3, 0.876 QWK) and a weaker Transformer teacher (Swin-Base, 0.830 QWK) are complementary, yet the Transformer's logits can still mislead the student. We therefore propose OrthKD, a selective-trust distillation framework that transfers full supervision from the strong CNN, uses feature-only distillation from the weak ViT, and enforces orthogonality between teacher-specific student projections to encourage complementary rather than redundant evidence. This design preserves local lesion precision, injects global structural context, and improves robustness to distribution shift. On 132,049 retinal images, a 5.4M-parameter MobileNetV3 student reaches 0.885 QWK on EyePACS and improves zero-shot Messidor-2 performance from 0.507 to 0.728 QWK, while also achieving strong referral AUC and calibration. These results show that selectively distilling heterogeneous teachers can enable practical DR screening on resource-constrained devices.
\end{abstract}

\section{Introduction}

\subsection{Technical Challenges in Medical AI Deployment}
Deep learning models now rival ophthalmologists in diabetic retinopathy (DR) screening~\cite{gulshan2016development,zhou2021review}, yet clinical impact remains limited. Haq et al.~\cite{haq2024computationally} analyzed 84 DR AI papers, noting that neglected computational complexity is a major shortcoming. While large models (EfficientNet-B7, ResNet-152, ViT-L) may achieve relatively high performance, they require GPU infrastructure and cloud connectivity that are difficult for primary healthcare settings to access. Meanwhile, 537 million high-risk individuals globally still lack screening accessibility.~\cite{teo2021global}.

Edge-friendly architectures such as MobileNetV3, EfficientNet-Lite, and ShuffleNetV2 support CPU inference~\cite{howard2019searching,zhang2018shufflenet}, but they often lose sensitivity near adjacent classification boundaries, precisely where annotator agreement is lowest (Cohen's kappa = 0.61--0.68)~\cite{krause2018grader}. This degradation is especially problematic for DR screening because (1) the task is ordinal, with disease severity progressing monotonically from grade 0 to 4; (2) the clinically important referable-DR threshold is grade $\geq 2$; and (3) missed positives at this threshold delay treatment for vision-threatening disease. Our goal is therefore not only to preserve high diagnostic performance, but also to make lightweight models generalize reliably across real clinical sites.

\subsection{Limitations of Standard Knowledge Distillation}
Knowledge distillation (KD)~\cite{hinton2015distilling} is a natural route to deployment because it compresses teacher knowledge into lightweight student models. However, standard KD assumes that teacher predictions are sufficiently reliable to be transferred directly. This assumption breaks in heterogeneous medical ensembles, where different architectures may be complementary but not equally trustworthy. In DR screening, a Transformer can capture useful global retinal context, yet still produce noisier logits than a CNN. Directly fusing their predictions therefore mixes complementary information with harmful decision noise.

This failure mode is particularly consequential in healthcare deployment. Real-world screening sites differ in cameras, acquisition protocols, and patient populations, so compressed models must remain reliable under distribution shift. Moreover, clinical decision-making is threshold-driven (e.g., referable DR defined by grade $\geq 2$), where both missed positives and overconfident errors are costly. It is therefore insufficient for distillation to improve only in-domain accuracy: a deployment-ready student must preserve grading fidelity, referral safety, and confidence reliability.

\subsection{The Expertise Asymmetry Challenge}
Our empirical analysis reveals three defining properties of the expertise-asymmetry regime.

First, the Transformer teacher (Swin-Base, 0.830 QWK) significantly lags behind both the CNN teacher (EfficientNet-B3, 0.876 QWK) and the student baseline (0.853 QWK). Such asymmetry is common in medical imaging, where CNN inductive bias often remains advantageous under limited data.

Second, weak teachers can still provide complementary knowledge. Despite its lower overall accuracy, the Transformer correctly classified 2.6\% of gold samples that both the CNN and student misclassified. These cases typically depend on global structural context rather than only local lesions.

Third, naive logit fusion is harmful. On disputed samples, dual-teacher logit distillation achieves only 73\% accuracy, yielding a 1:6 harm-to-benefit ratio. In other words, the weak teacher's errors dominate the rare cases where its predictions help.

These findings motivate a selective-trust strategy: preserve the weak teacher's useful feature-level context while blocking its unreliable decision-level supervision.

\subsection{OrthKD: Our Solution}
We propose OrthKD (Orthogonal Feature Knowledge Distillation), which addresses expertise asymmetry through two coupled ideas. First, we use \textit{feature-only distillation from the weak teacher}: the CNN supplies reliable decision-level supervision, while the ViT contributes only global contextual features. Second, we impose \textit{orthogonal constraints} on the teacher-specific student projections so that the two transferred signals remain complementary rather than collapsing to the same shortcuts. Together, these choices explain why each branch in Figure~\ref{fig:framework} is needed and why the overall pipeline is effective.

\subsection{Contributions}
This paper makes three main contributions. A minimal public reference implementation is available online.\footnote{Project page: \url{https://github.com/ScottBlizzard/orthkd-dr-screening}.}
\begin{itemize}[leftmargin=*,itemsep=2pt,topsep=2pt]
    \item \textit{Problem formulation and insight.} We identify \emph{expertise asymmetry} in heterogeneous KD for medical imaging: weak teachers may be decision-level unreliable yet still contain complementary feature-level knowledge.
    \item \textit{Method.} We introduce OrthKD, a selective-trust distillation framework that combines full supervision from a strong CNN, feature-only transfer from a weak ViT, and orthogonal regularization between teacher-specific student projections.
    \item \textit{Evaluation and deployment value.} With a 5.4M-parameter MobileNetV3 student, OrthKD achieves strong in-domain QWK/AUC/ECE and improves zero-shot Messidor-2 performance from 0.507 to 0.728 QWK, supporting practical DR screening on resource-constrained devices.
\end{itemize}

\section{Related Work}

Knowledge distillation (KD)~\cite{hinton2015distilling,li2024kd} transfers knowledge by minimizing KL divergence between teacher and student logits. Multi-teacher KD~\cite{you2017learning,mirzadeh2020improved} typically assumes that all teachers provide equivalent high-quality guidance. Recent heterogeneous distillation works~\cite{passalis2020heterogeneous} address the architecture mismatch (CNN to Transformer), but still assume comparable teacher performance. In contrast, our work identifies and addresses the Expertise Asymmetry scenario where a weak teacher (e.g., underperforming ViT) is leveraged for its unique complementary features despite its unreliable decisions.

Feature-level distillation~\cite{romero2014fitnets} and attention transfer~\cite{zagoruyko2016paying} enable students to learn richer intermediate representations. Diversity-promoting multi-teacher methods are also relevant in spirit, but they typically assume symmetric teacher reliability: they either aggregate all teacher predictions or encourage diversity without distinguishing which teacher signals should be trusted. In expertise asymmetry, this is insufficient because the main problem is not only redundancy, but also harmful decision-level supervision from the weak teacher. OrthKD addresses both issues jointly by blocking weak-teacher logits and regularizing the two transferred feature streams to remain complementary.

Deploying AI for DR screening in resource-constrained primary care settings requires both efficiency and cross-dataset robustness~\cite{cleland2023artificial,zhou2021review}. While standard compression techniques (quantization, pruning) yield lightweight models~\cite{howard2019searching,han2015deep}, they often overfit to the training distribution and fail on unseen clinical populations. Our work shows that combining dual-teacher KD with explicit diversity control can improve out-of-distribution robustness without sacrificing deployment efficiency.

Traditional methods often enforce strict spatial alignment between student and teacher attention maps~\cite{zagoruyko2016paying}. However, in our asymmetric expertise setup, weak teacher attention captures global context but lacks the local fineness of CNNs; strict alignment forces students to learn these shortcomings, thereby capping performance below the ceiling of strong teachers. In contrast, OrthKD's orthogonal constraints foster complementarity: students can retain CNN precision while selectively absorbing Transformer-based global context.

\section{Methodology}

\begin{figure*}[t]
    \centering
    \includegraphics[width=0.95\textwidth]{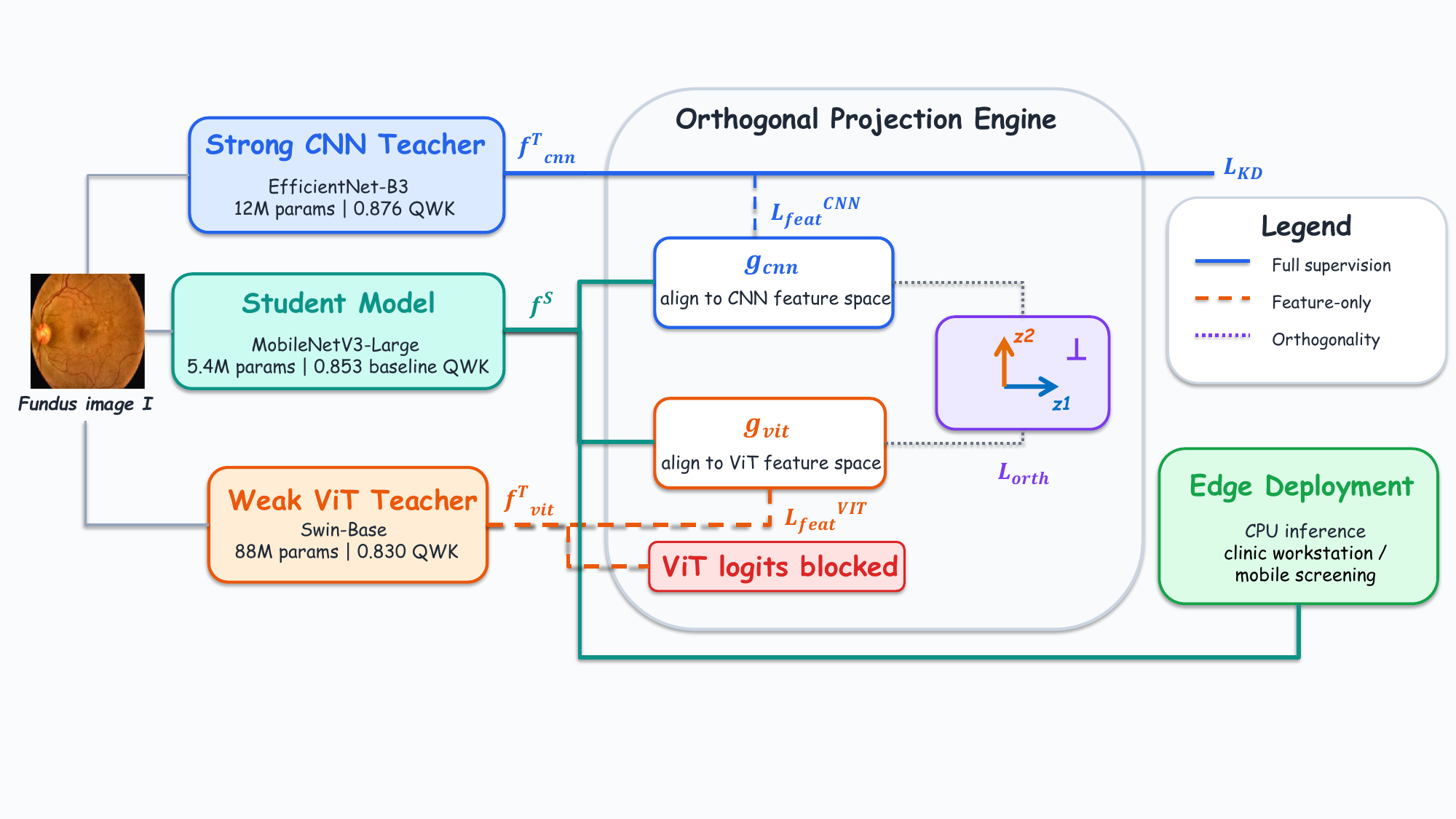}
    \caption{\textbf{OrthKD Framework.}
    Given a fundus image, three networks extract features in parallel: a strong CNN teacher (EfficientNet-B3, 0.876 QWK), a lightweight student (MobileNetV3), and a weak ViT teacher (Swin-Base, 0.830 QWK).
    The student learns through two projection heads: $g_{cnn}$ aligns with CNN features and receives \textit{full supervision} (logits + features), while $g_{vit}$ aligns with ViT features but receives \textit{feature-only supervision} (ViT logits are blocked).
    The orthogonality constraint $\mathcal{L}_{orth}$ enforces complementary projected features, discouraging redundancy. This role separation is crucial: CNN logits anchor the decision boundary, ViT features supply complementary global structure, and orthogonality prevents redundant transfer.}
    \label{fig:framework}
\end{figure*}

OrthKD follows a selective-trust design. The strong CNN teacher provides both logits and features because it is the most reliable classifier, while the weak ViT teacher provides only features because its value lies in global structural context rather than decision-level supervision. The student learns one projection for each teacher, and an orthogonality constraint prevents these branches from collapsing to the same evidence. Figure~\ref{fig:framework} summarizes the full pipeline.

\paragraph{Feature representation.}
Let $z$ denote pre-softmax logits and let $f$ denote the intermediate representation used for distillation. In all networks, we take $f$ from the final backbone stage before the classification head. If the native representation is spatial, we apply global average pooling to obtain a single vector. This yields compact features $f^S, f^{CNN}, f^{ViT} \in \mathbb{R}^{d}$, which are compatible with lightweight deployment and stable feature alignment.

\subsection{Problem Formulation}

We address five-grade DR severity classification $y \in \{0,1,2,3,4\}$ with ordinal structure following the ICDR scale~\cite{wilkinson2003proposed}. We select EfficientNet-B3 ($\mathcal{T}_{CNN}$, 0.876 QWK) as the strong CNN teacher and Swin-Base ($\mathcal{T}_{ViT}$, 0.830 QWK) as the Transformer teacher---the most competitive among tested ViT variants. This asymmetry in expertise (CNN $>$ ViT) is not merely an experimental anomaly, but a common characteristic in medical imaging: unlike natural vision where ViTs thrive on massive datasets (e.g., JFT-300M), medical domains are data-scarce relative to model capacity. Under this regime, CNNs' inductive biases (e.g., translation invariance) provide a persistent advantage.

The student is MobileNetV3-Large (5.4M params), achieving 0.853 QWK when trained independently. Standard dual-teacher distillation that combines logits from both teachers leads to performance degradation---the weaker Transformer introduces erroneous guidance. Our analysis shows only 73\% accuracy on controversial samples, with a harm-to-benefit ratio of 1:6 (Section~\ref{sec:analysis}).
\label{sec:analysis}

\subsection{Orthogonal Feature Distillation}
Existing multi-teacher KD pipelines often treat all teachers symmetrically. In our setting, this is suboptimal: symmetric logit fusion propagates weak-teacher mistakes, while symmetric feature matching can still learn redundant cues. We therefore assign asymmetric roles to the two teachers. Although the weak Transformer has lower classification accuracy, it captures useful global context and correctly classifies 2.6\% of samples where both the CNN and the student fail. We transfer this knowledge only at the feature level:
\begin{itemize}[leftmargin=*,itemsep=1pt,topsep=2pt]
    \item \textit{CNN Teacher (Strong):} Full transfer---logits + features.
    \item \textit{ViT Teacher (Weak):} Feature-only transfer---no logit supervision.
\end{itemize}
The two projection heads are lightweight distillation modules used only during training; the deployed model remains the plain MobileNetV3 classifier, so inference cost is unchanged.

\begin{figure}[t]
    \centering
    \includegraphics[width=\columnwidth]{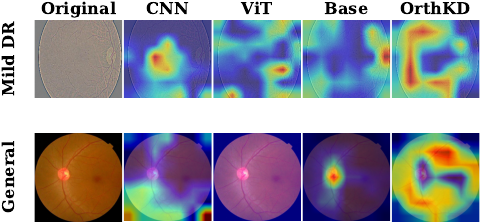}
    \caption{\textbf{Grad-CAM visualization.}
    CNN focuses on local textures (e.g., optic disc), while ViT exhibits diffuse attention capturing global context.
    The baseline student overfits to salient landmarks, whereas the OrthKD student localizes lesions and better captures vascular structures.}
    \label{fig:gradcam}
\end{figure}

\subsection{Qualitative Analysis}
Figure~\ref{fig:gradcam} visualizes Grad-CAM maps for two representative samples.

In the lower row, the CNN teacher and baseline student primarily focus on the optic disc, a highly prominent local landmark. This indicates that without proper guidance, the lightweight student tends to ``learn shortcuts''---classifying based on simple landmarks, especially bright areas, rather than understanding the retinal structure. In contrast, the ViT teacher has a more global receptive field and focuses on the entire retinal background to confirm the absence of lesions.

More importantly, the OrthKD student model (column 5) exhibits a unique "hybrid" attention pattern: while retaining the sharpness typical of CNN structures (derived from the convolutional architecture), its activation regions are significantly larger compared to the baseline. Specifically, it successfully tracks the arc-shaped trajectory of blood vessels extending from the optic disc, a behavior absent in the baseline but consistent with the global perception of ViT.

This validates our hypothesis: by imposing orthogonality between student projection features, OrthKD avoids model collapse to solutions that only mimic the easiest features (optic disc); instead, it forces the student to explore orthogonal, complementary subspaces, thereby learning richer semantic information (such as vascular topology), achieving a combination of CNN precision and ViT breadth.

Let $f^S, f^{CNN}, f^{ViT} \in \mathbb{R}^d$ be student and teacher feature representations. The Orthogonal Feature KD loss combines three components: (1) logit KD from the CNN teacher,
\begin{equation}
\mathcal{L}_{\mathrm{KD}}^{CNN} = \tau^2 \cdot \mathrm{KL}\left(\sigma(z^S/\tau), \sigma(z^{CNN}/\tau)\right)
\end{equation}
where $\sigma(\cdot)$ is the softmax and $\tau$ is the distillation temperature; (2) feature alignment with both teachers,
\begin{equation}
\begin{aligned}
\mathcal{L}_{\text{feat}}^{CNN} &= \|g_{cnn}(f^S) - f^{CNN}\|_2^2, \\
\mathcal{L}_{\text{feat}}^{ViT} &= 1 - \cos(g_{vit}(f^S), f^{ViT})
\end{aligned}
\end{equation}
where $g_{cnn}(\cdot)$ and $g_{vit}(\cdot)$ are learnable projection layers adapting student features to each teacher's latent space; and (3) the orthogonality constraint,
\begin{equation}
\mathcal{L}_{\text{orth}} = \cos^2(g_{cnn}(f^S), g_{vit}(f^S))
\label{eq:orth}
\end{equation}
 This penalizes the correlation between the student's two projection heads. We adopt the squared cosine similarity as it provides smoother gradients near zero compared to absolute-value formulations, facilitating more stable optimization of distinct feature subspaces for satisfying both teachers simultaneously.

\subsection{Training Objective}
The complete OrthKD loss combines classification, knowledge distillation, and orthogonality:
\begin{equation}
\mathcal{L}_{\mathrm{total}} = \alpha_1 \mathcal{L}_{\mathrm{CE}} + \alpha_2 \mathcal{L}_{\mathrm{KD}}^{CNN} + \alpha_3 \mathcal{L}_{\text{feat}}^{CNN} + \alpha_4 \mathcal{L}_{\text{feat}}^{ViT} + \alpha_5 \mathcal{L}_{\text{orth}}
\end{equation}
where $\mathcal{L}_{\mathrm{CE}}$ is cross-entropy with ground truth, $\mathcal{L}_{\mathrm{KD}}^{CNN}$ is logit distillation from the CNN teacher, $\mathcal{L}_{\text{feat}}$ are feature alignment losses, and $\mathcal{L}_{\text{orth}}$ enforces orthogonality. We use $(\alpha_1, \alpha_2, \alpha_3, \alpha_4, \alpha_5) = (0.2, 0.4, 0.2, 0.1, 0.1)$: the larger weights are assigned to ground-truth supervision and the reliable CNN logit signal, while ViT feature transfer and orthogonality act as auxiliary regularizers. Importantly, these weights are fixed across all variants and datasets to avoid over-tuning to any particular test condition.

\subsection{Why Orthogonality Improves OOD Robustness}
A significant observation in our experiments is that orthogonal constraints have almost negligible impact on in-domain performance but lead to significant improvements in out-of-domain performance. Specifically, removing orthogonality only causes a slight degradation on the training distribution (Table~\ref{tab:ablation}), but leads to a noticeable decline on external clinical datasets such as Messidor-2. We attribute this to three complementary mechanisms.

First, orthogonality prevents \textit{redundancy collapse}. Without explicit decorrelation, the two projection heads $g_{cnn}$ and $g_{vit}$ tend to converge to similar representations---encoding the ``easiest'' discriminative features (e.g., the optic disc). Minimizing $\cos^2(g_{cnn}(f^S), g_{vit}(f^S))$ forces them into different subspaces, allowing the student to simultaneously capture local lesion evidence from the CNN and global structural context from the ViT.

Second, orthogonality \textit{mitigates shortcut learning}. Lightweight students tend to exploit spurious correlations associated with salient landmarks. The weaker Transformer, despite lower accuracy, focuses on background structures less dependent on domain-specific features. Orthogonality constraints force the student to retain this ``background-aware channel,'' regularizing against shortcut reliance.

Third, orthogonality provides \textit{diversity as a buffer against distribution shift}. When training and external distributions differ (e.g., due to imaging equipment or population characteristics), redundant features degrade simultaneously. Orthogonal subspaces provide decorrelated evidence streams; when one stream degrades, the other may remain stable, yielding more robust predictions.

In summary, orthogonality $\to$ complementary feature subspaces $\to$ reduced overfitting to domain-specific shortcuts $\to$ stronger OOD generalization. This causal chain explains why orthogonality is ``silent'' within a single domain but crucial in real-world clinical deployments.

\subsection{Deployment Considerations}
We select MobileNetV3-Large (5.4M parameters) as the student architecture for its deployment-friendly design:
\begin{itemize}[leftmargin=*,itemsep=2pt]
\item \textit{Parameter efficiency:} 5.4M params (vs. 12M B3, 88M Swin).
\item \textit{Mobile-optimized:} Inverted residuals and hard-swish activations.
\item \textit{Edge deployment:} Designed for CPU inference without GPU requirements.
\end{itemize}
This makes DR screening possible in resource-constrained primary healthcare settings that lack GPU infrastructure.

Beyond efficiency, clinical deployment also requires models to perform stably near the referral threshold (grade $\geq$ 2) and be robust to variations in image quality. Our strong performance on Messidor-2 (0.728 QWK)---a dataset that likely exhibits greater domain shift and variability than the high-quality IDRiD dataset---validates the potential of OrthKD in primary care settings. The OrthKD student model inherits the clinical accuracy of CNNs while incorporating the complementary perspective of Transformers, resulting in more stable predictions at the decision boundary where clinical risk is highest. In a typical primary care triage process, this model can assist non-specialist screening personnel in flagging referral cases for ophthalmologist review, reducing workload while maintaining a safety margin.
\section{Experiments}

\subsection{Datasets and Experimental Protocol}
We combine two public DR screening datasets with data augmentation:
\begin{itemize}[leftmargin=*,itemsep=1pt,topsep=1pt]
    \item \textit{EyePACS} (Kaggle Diabetic Retinopathy Detection Challenge, 2015): 126,584 fundus images from California screening sites, labeled with 5-grade ICDR scales~\cite{wilkinson2003proposed}.
    \item \textit{APTOS} (Asia Pacific Tele-Ophthalmology Society Challenge, 2019): 5,465 images from Indian population, using the same 5-grade labeling~\cite{aptos2019}.
\end{itemize}
Total training pool: 132,049 base images, expanded via comprehensive data augmentation. We use a stratified split: 80\% train, 10\% validation, 10\% test. Teachers and student trained on the combined dataset.

To assess cross-dataset generalization under distribution shift:
\begin{itemize}[leftmargin=*,itemsep=1pt,topsep=1pt]
    \item \textit{Messidor-2} (European population): N=1,728, macula-centered images with expert adjudication. Not seen during training.
    \item \textit{IDRiD} (Indian Diabetic Retinopathy Image Dataset)~\cite{porwal2018indian}: N=516, high-resolution fundus images from an Indian clinical setting with expert-verified 5-level grading. This dataset represents real-world clinical validation with different population demographics and imaging equipment.
\end{itemize}

The black borders were removed using circular cropping, and CLAHE (clip limit=2.0) was used for illumination normalization, followed by scaling to 512$\times$512. No specific parameter tuning was performed on the external dataset (zero-shot evaluation).

The main reporting metric is quadratic weighted kappa (QWK), because DR grading is a five-level ordinal problem and QWK directly measures agreement on disease severity. We additionally report ROC-AUC for referable DR (grade $\geq 2$), since screening systems are ultimately used to separate referral from non-referral cases and high AUC reduces the risk of missed high-risk patients across operating points. We also report Expected Calibration Error (ECE, 15 bins), because a deployable screening model must assign confidence scores that are reliable enough for triage and manual review. Together, QWK, AUC, and ECE capture grading fidelity, screening safety, and confidence reliability, respectively. For external validation sets (Messidor-2 and IDRiD), we focus on QWK because zero-shot evaluation should not depend on target-domain threshold tuning or recalibration.

\subsection{Implementation Details}
We first train EfficientNet-B3, Swin-Base, and the MobileNetV3 baseline independently on the same 80/10/10 split of the combined training pool, and then train distillation variants while keeping the student architecture fixed (MobileNetV3-Large). Images are circularly cropped, normalized with CLAHE (clip limit 2.0), resized to 512$\times$512, and augmented with random flips, small rotations, color jitter, and Cutout. All models are optimized with AdamW (learning rate $3\times10^{-4}$, weight decay $10^{-4}$), a cosine schedule, batch size 32, and 50 training epochs. For fair comparison, we use the same preprocessing, optimizer family, validation protocol, and fixed loss weights across student variants, and we avoid any tuning on Messidor-2 or IDRiD. All tables report single-run results from one training run per configuration; broader multi-seed statistical analysis is left for future work and discussed in Limitations.

We compare our method against six configurations: (1) \textit{EfficientNet-B3}, the CNN teacher (12M params, 0.876 QWK); (2) \textit{Swin-Base}, the ViT teacher (88M params, 0.830 QWK); (3) \textit{MobileNetV3}, the student baseline (5.4M params, 0.853 QWK); (4) \textit{B3-Only KD}, standard single-teacher distillation using only CNN cues; (5) \textit{CW Logit Fusion}, a multi-teacher baseline using confidence-weighted logits; (6) \textit{OrthKD (Ours)}, the proposed orthogonal feature distillation from heterogeneous teachers. These baselines isolate three questions: whether a strong single teacher is sufficient, whether naive multi-teacher decision fusion is sufficient, and whether dual-teacher feature transfer still requires explicit diversity control. We do not include co-training diversity methods as direct baselines because they change the setting from heterogeneous distillation to ensemble training and still do not solve weak-teacher logit contamination.

\section{Results}

\subsection{Main Results: Clinical Performance and Safety}

\begin{table*}[!t]
\centering
\small
\setlength{\tabcolsep}{5pt}
\begin{tabular}{@{}lcccccc@{}}
\toprule
\multirow{2}{*}{\textbf{Method}} & \multicolumn{3}{c}{\textbf{EyePACS (In-Domain)}} & \multicolumn{3}{c}{\textbf{APTOS (In-Domain)}} \\
\cmidrule(lr){2-4} \cmidrule(l){5-7}
 & \textbf{QWK} $\uparrow$ & \textbf{AUC} $\uparrow$ & \textbf{ECE} $\downarrow$ & \textbf{QWK} $\uparrow$ & \textbf{AUC} $\uparrow$ & \textbf{ECE} $\downarrow$ \\
\midrule
Swin-Base (ViT Teacher) & .830 & .935 & \textbf{.005}$^\dagger$ & .918 & .980 & .038 \\
EffNet-B3 (CNN Teacher) & .876 & .965 & .045 & .941 & .993 & .032 \\
MobileNetV3 (Baseline) & .853 & .947 & .037 & .934 & .987 & .030 \\
\midrule
B3-Only KD & .878 & .965 & .027 & .942 & .994 & .040 \\
w/o Orthogonality & .883 & .965 & .035 & .950 & .995 & .031 \\
\textbf{Ours (OrthKD)} & \textbf{.885} & \textbf{.966} & .023 & \textbf{.952} & \textbf{.995} & \textbf{.018} \\
\bottomrule
\multicolumn{7}{l}{\scriptsize $^\dagger$ Swin's low ECE on EyePACS may reflect underfitting (overly low-confidence predictions) rather than well-calibrated high-accuracy predictions.}
\end{tabular}
\caption{Comprehensive in-domain performance on QWK, AUC, and ECE. OrthKD achieves the best overall balance across grading fidelity, referral safety, and calibration.}
\label{tab:main_results}
\end{table*}

In actual clinical diagnosis, the three most frequently used metrics are grading accuracy (QWK), screening safety (AUC for referral vs. non-referral), and reliability (Expected Calibration Error, ECE).

Table~\ref{tab:main_results} provides a comprehensive evaluation of the model's performance based on the three metrics mentioned above.

OrthKD achieved the highest classification accuracy on two domain datasets (EyePACS: 0.885 QWK; APTOS: 0.952). More importantly, it performed comparably to or even better than strong teacher models in terms of screening safety, achieving the highest AUC (0.966 / 0.995). This demonstrates that model compression does not weaken its ability to identify referral cases, which is a core requirement for clinical safety.

Additionally, OrthKD demonstrates superior calibration (ECE = 0.023 on EyePACS), significantly outperforming the overconfident CNN teacher (0.045) and the baseline student (0.037). Comparing ``w/o Orthogonality'' and ``OrthKD,'' although the QWK scores are similar, we observe a significant decrease in ECE ($0.035 \to 0.023$), indicating that the orthogonality constraint effectively regularizes the student's confidence, preventing it from collapsing into a redundant or overconfident feature space.

\textit{Note on Swin Transformer:} Swin-Base achieves an ECE close to 0 (0.005) on EyePACS, but its accuracy is lower (0.830 QWK). This suggests that a low ECE might reflect underfitting (overall low confidence) rather than good calibration at high accuracy. In contrast, OrthKD achieves a low ECE while maintaining high accuracy, demonstrating better reliability.

\subsection{Ablation Study}
\label{sec:ablation}
Table~\ref{tab:ablation} verifies the contribution of each component. Removing the ViT features (B3-Only) results in a 0.9\% performance drop, indicating that the weak teacher indeed provides valuable complementary knowledge. Orthogonal constraints have only a marginal impact within the domain (approximately 0), while confidence-weighted logit fusion performs significantly worse (-1.6\%), further confirming that using weak teacher logits should be avoided.

\begin{table}[ht]
\centering
\small
\begin{tabular}{@{}lccc@{}}
\toprule
\textbf{Variant} & \textbf{APTOS} & \textbf{EyePACS} & \textbf{$\Delta$} \\
\midrule
Full OrthKD & .952 & .885 & --- \\
	w/o ViT Features (B3-Only) & .942 & .878 & -0.9\% \\
	w/o Orthogonality & .950 & .883 & $\approx$0 \\
	CW Logit Fusion & .937 & .869 & -1.6\% \\
\bottomrule
\end{tabular}
\caption{Ablation study. Each row removes one component from the full OrthKD.}
\label{tab:ablation}
\end{table}

\subsection{Cross-Dataset Generalization}

Our core hypothesis is that orthogonality primarily acts as a diversity regularizer under distribution shift, rather than improving in-domain performance. Table~\ref{tab:crossdataset} supports this: although orthogonality provides only a small gain on the training set (see Table~\ref{tab:ablation}), it brings a +7.4 percentage point improvement on Messidor-2 (0.728 vs 0.654), significantly narrowing the gap with the strong teacher model.

In external validation, we emphasized zero-shot generalization for realism, and therefore did not employ dataset-specific threshold tuning or post-processing calibration. Consequently, we used the Quadratic Weighted Kappa (QWK) on Messidor-2 and IDRiD as the primary metric, as it directly measures ordinal classification agreement and is insensitive to operating point selection during deployment. Within the domain, we additionally report referral AUC and ECE to characterize screening safety and reliability under matching distributions.

\begin{table}[ht]
\centering
\small
\begin{tabular}{@{}lcc@{}}
\toprule
\textbf{Method} & \textbf{Messidor-2} & \textbf{IDRiD} \\
\midrule
B3 (CNN Teacher) & .785 & .749 \\
Swin (ViT Teacher) & .453 & .636 \\
MobileNetV3 (Baseline) & .507 & .516 \\
\midrule
B3-Only KD & .587 & .701 \\
No Orthogonality & .654 & .724 \\
CW Logit Fusion & .659 & .728 \\
\textbf{Ours (OrthKD)} & \textbf{.728} & \textbf{.743} \\
\bottomrule
\end{tabular}
\caption{External validation under zero-shot transfer. Comparison of all methods on unseen clinical datasets.}
\label{tab:crossdataset}
\end{table}

\begin{figure*}[t]
    \centering
    \includegraphics[width=\textwidth]{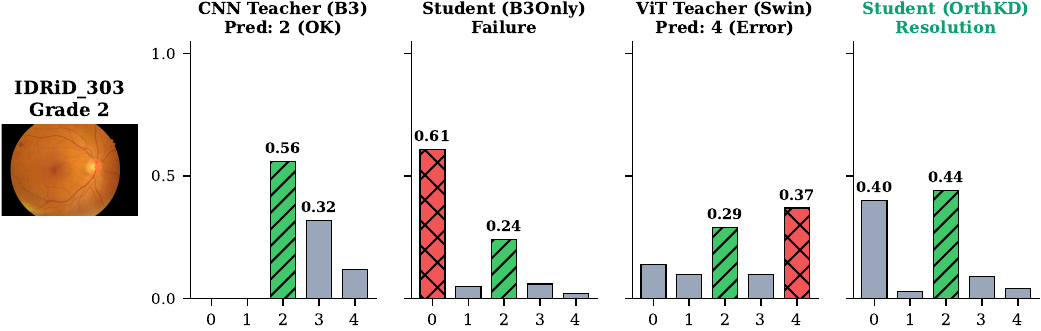}
    \caption{\textbf{Ambiguity resolution case study.}
    On IDRiD\_303 (Grade 2), the baseline student (B3-Only) overfits to local textures and predicts Grade 0, while the CNN teacher predicts Grade 2.
    Although the weak ViT's logits are unreliable, its features provide global structural context; OrthKD leverages this feature-level complementarity via orthogonality. Hatching preserves the key cues under grayscale printing.}
    \label{fig:ambiguity}
\end{figure*}

OrthKD outperforms all other student variants across both external datasets, a pattern also illustrated in Figure~\ref{fig:ambiguity}. Compared to the single-teacher B3-Only model, it improves by 14.1 percentage points on Messidor-2 and 4.2 percentage points on IDRiD. This suggests that capturing the ``global intelligence'' of weak Transformers is more effective than simply mimicking the most accurate teacher. Notably, feature transfer from Swin (``No Orthogonality'') already outperforms B3-Only KD on Messidor-2 (0.654 vs. 0.587), supporting our claim that weak-teacher features contain useful complementary evidence when decoupled from unreliable logits.

\section{Discussion}
\subsection{Key Insight: Orthogonality Enables Generalization}
Orthogonal constraints are the main driver of OrthKD's generalization gains. While their in-domain effect is small, they improve Messidor-2 by +7.4 percentage points over the non-orthogonal variant, indicating that their primary role is to prevent redundant, shortcut-prone representations rather than to inflate training-set accuracy.

\subsection{AI \& Health Perspective: Screening Safety and Deployment}
In a screening workflow, model utility depends not only on ordinal grading accuracy, but also on safe behavior around referable DR thresholds and robustness to site-to-site variability. This is why we report AUC and calibration alongside QWK and emphasize zero-shot external validation without target-domain tuning.

\subsection{Why Naive Alternatives Fail}
Three progressively stronger alternatives clarify why the full OrthKD pipeline is needed. Confidence-weighted logit fusion still lets weak-teacher errors dominate disputed cases (73\% accuracy), showing that selective supervision is necessary. Feature-only transfer from the ViT improves in-domain performance, confirming that the weak teacher does contain useful contextual evidence, but it remains vulnerable to redundant representations. Adding orthogonality resolves this redundancy and yields the strongest external generalization (+7.4 pp on Messidor-2 over the non-orthogonal variant). This progression explains why OrthKD requires both asymmetric teacher roles and explicit diversity control.

\subsection{Why Single-Teacher KD Generalizes Poorly}
Our strongest teacher B3 transfers only part of its external robustness under single-teacher KD (0.587 vs 0.785 on Messidor-2). We attribute this to the combination of a student capacity bottleneck and the absence of complementary supervision, which explains why OrthKD substantially outperforms B3-Only.

\subsection{Implications Beyond DR}
Although this paper focuses on DR grading, the same selective-trust principle should extend to other medical tasks where heterogeneous teachers are complementary but unequally reliable.

\subsection{Limitations}
This work has three main limitations. First, we report single-run results without multi-seed confidence intervals, so broader statistical validation remains an important next step. Second, regarding \textit{scalability}, we only study a two-teacher setting; extending orthogonality to larger teacher sets deserves further study. Third, regarding \textit{theoretical understanding}, why orthogonality strongly improves out-of-domain robustness while only mildly affecting in-domain performance remains an empirical observation rather than a formal guarantee.

\section{Conclusion}

OrthKD addresses a practical deployment problem in medical AI: how to distill heterogeneous teachers when they are complementary but not equally reliable. Our method assigns asymmetric roles to the two teachers: the strong CNN provides full decision-level supervision, the weak ViT contributes feature-level global context only, and orthogonality keeps the transferred signals complementary rather than redundant.

This design yields a 5.4M-parameter MobileNetV3 student that substantially improves zero-shot generalization on unseen clinical data: OrthKD raises Messidor-2 performance from 0.507 to 0.728 QWK while preserving strong in-domain QWK, referral AUC, and calibration. More broadly, our results suggest that heterogeneous distillation for medical imaging should be organized around \emph{signal reliability} rather than teacher strength alone, motivating extensions to larger teacher sets and broader multi-site evaluation.

\clearpage
\begingroup
\small
\bibliographystyle{named}
\bibliography{ijcai26}
\endgroup

\end{document}